\documentclass{Interspeech2024}

\usepackage{graphicx}
\usepackage{subcaption}
\usepackage{diagbox} 
\usepackage{cite}

\interspeechcameraready

\title{From Text to Emotion: Unveiling the Emotion Annotation Capabilities of LLMs}

\name[affiliation={1}]{Minxue}{Niu}
\name[affiliation={2}]{Mimansa}{Jaiswal}
\name[affiliation={1}]{Emily}{Mower Provost}

\address{
  $^1$University of Michigan, USA
  $^2$Independent Researcher
  }
\email{sandymn@umich.edu, mimansa.jaiswal@gmail.com, emilykmp@umich.edu}


\begin{document}
\maketitle

\begin{abstract}
Training emotion recognition models has relied heavily on human annotated data, which present diversity, quality, and cost challenges. In this paper, we explore the potential of Large Language Models (LLMs), specifically GPT-4, in automating or assisting emotion annotation. We compare GPT-4 with supervised models and/or humans in three aspects: agreement with human annotations, alignment with human perception, and impact on model training. We find that common metrics that use aggregated human annotations as ground truth can underestimate GPT-4's performance, and our human evaluation experiment reveals a consistent preference for GPT-4 annotations over humans across multiple datasets and evaluators. Further, we investigate the impact of using GPT-4 as an annotation filtering process to improve model training. Together, our findings highlight the great potential of LLMs in emotion annotation tasks and underscore the need for refined evaluation methodologies.

\end{abstract}

\section{Introduction}
Understanding human emotions from written or spoken language is crucial is a key part of studying how computers can interact with us more like humans do. The field has attracted significant research efforts, ranging from word-level analysis~\cite{mohammad2018obtaining, hung2023beyond} to building sophisticated neural networks~\cite{alswaidan2020survey, deng2021survey}. Currently, many models demonstrate impressive capabilities in recognizing various human emotions.

The training of emotion models has relied heavily on datasets with human annotations. However, obtaining emotion annotations is challenging due to the rich, ambiguous and subjective nature of emotions~\cite{wu2023estimating, tran2022multimodal, devillers2005challenges}. 
The first challenge is to identify the emotion theory that will motivate a particular labeling schema. 
Common theories include basic emotion theory~\cite{ekman1999basic}, assigning one or more predefined emotion classes to each sample (categorical labels), and the emotion circumplex theory~\cite{russell1980circumplex}, rating each sample on continuous scales, such as valence and arousal to reflect the emotion's positivity and intensity (dimensional labels). The process of collecting human annotations involves multiple annotators per sample to accommodate subjective interpretations and possible quality issues, with the final label often determined through aggregation methods like majority voting~\cite{rosenthal2019SemEval} or averaging~\cite{Buechel2017-zn}. Given the large scale of modern datasets, such annotation processes can be both costly and time-consuming. Moreover, 
the complexity of the label space and the difficulty of quality control 
further add to the challenges.

Recently, the progress in LLMs brings new alternatives. With remarkable proficiency in language modeling across a wide range of scenarios, LLMs show emerging common sense reasoning capability~\cite{huang2022towards}: they can answer a wide range of natural language reasoning questions through zero- or few- shot prompting, matching or even outperforming supervised models~\cite{Laskar2023-gv, Brown2020-ni, Wei2021-xo}. What's more, LLMs display an understanding of human emotion and can respond  differently to emotional content~\cite{Zhao2023-wy, Huang2023-pn}. 
This has inspired research into leveraging LLMs as emotion models to aid emotion annotation processes.  

In this work, we comprehensively assess GPT-4's potential to perform emotion annotations in a zero-shot manner. We first measure its emotion recognition performance and find that it performs comparably to established supervised models as baselines, using human annotations as the ground truth. We then reflect on the differences between GPT-perception and human-perception and evaluate how those differences are perceived by a separate set of human evaluators. Surprisingly, we find that human evaluators consistently prefer the GPT-4 annotations over human annotations. These findings raise important open questions about the suitability of conventional ``ground truth'' concepts and evaluation practices, especially as models begin to approach human-level performance.
Further, we analyze how label formats (categorical vs. dimensional) affect GPT-4's performance, and we explore the feasibility of applying GPT-4 as a quality checker for existing annotations. We demonstrate that GPT-4 can identify potentially low-quality annotations and help with curating a cleaner and more efficient training set.

In summary, our research reveals the great potential of utilizing LLMs for emotion annotation tasks, offers new insights into their capabilities across label formats, and highlights the challenges involved in their evaluation. We also release the GPT-4 annotations on the entire GoEmotions dataset, along with our code and prompts\footnote{\url{https://github.com/chailab-umich/GPT-4-Emotion-Annotation}}.

\section{Related Work}
\textbf{Affective capabilities of LLMs.}
Many evaluation studies have shown that LLMs are equipped with emotional intelligence: they are able to derive appraisals of given situations~\cite{Tak2023-fk}, identify the emotions and emotion causes in dialogues~\cite{Zhao2023-wy}, and respond with emotional support~\cite{Zhao2023-wy, Huang2023-pn, li2023emotionprompt}. Yet, they are generally found to be inferior to humans: a few works that developed benchmarks for assessing emotional intelligence consistently indicate a notable gap in complex emotion reasoning between state-of-the-art LLMs and human performance~\cite{Wang2023-ur, Huang2023-pn, Sabour2024-pa}. There have been a few works that evaluate GPT's zero- or few-shot capability of emotion recognition from text or speech input~\cite{Wake2023-rn, Feng2023-rh, Latif2023-wl, Zhang2023-gl}. However, the diversity of emotion label spaces are rarely discussed. Besides, existing works adopt evaluation criteria that rely on automatic metrics against human annotations as the ground truth. In this work, we show that such metrics can be biased and may undervalue GPT's effectiveness.

\textbf{LLMs as data annotators.} 
Despite their remarkable capabilities in various language understanding tasks~\cite{Brown2020-ni, Wei2021-xo, Sun2023-pp}, the high operational costs and impracticality of deployment on edge devices have focus efforts towards using LLMs to enhance annotation processes for training more compact models. 
GPT has been recognized for its potential in sample annotation~\cite{Gilardi2023-kv} and generation~\cite{Ding2023-xi, Thapa2023-vf}. In a closer look, LLMs especially excel at tasks with limited and well-defined label sets~\cite{Ding2023-xi}. 

\textbf{Prompting methods.}
It is widely acknowledged that LLMs are sensitive to the format and word choices in the prompts~\cite{loya2023exploring}, making prompts the key factor in the successful application of LLMs. There are two common ways of prompting~\cite{Liu2023-yz}: cloze prompts, which involve a fill-in-the-blank approach (e.g.,``I feel [X]. I finally got that promotion!"), and prefix prompts, where the model extends a given prompt (e.g., ```I finally got the promotion!' What is the speaker's emotion?"). Given GPT-4's pretraining on generation tasks, our study employs prefix prompts. There have been a lot of work exploring different techniques of prompting that could bring a significant improvement in the models' responses~\cite{Liu2023-yz, Binz2023-in, wei2022chain}. The efficiency of different prompting techniques is not the focus of this paper. We follow the common effective practices without dedicated prompt engineering (details in Section \ref{sec:prompt_method}).

\label{sec:prompt_literature}
\vspace{-5pt}
\section{Data}

We use four publicly available emotion recognition datasets for our analysis, encompassing a variety of label representations and diverse text domains (Table \ref{table:datasets}). Considering the substantial volume of these datasets, we first select a subset of 500 samples from each for GPT-4 annotation and subsequent analysis. 

\begin{table}[t]
  \caption{Dataset details. label: C-categorical, D-dimensional. The column ``Multi'' indicates whether it's a multilabel classification task. ``Indiv.'' indicates whether individual annotations on each sample are released.  }
  \label{table:datasets}
  \vspace{-7pt}
  \centering
  \begin{tabular}{@{}ccccc@{}}
\toprule
  \multicolumn{1}{c}{\textbf{Dataset}}       & \multicolumn{1}{c}{\textbf{Domain}} & \multicolumn{1}{c}{\textbf{Label (d)}} &\textbf{ Multi.} & \multicolumn{1}{c}{\textbf{Indiv.}} \\ \midrule
ISEAR      & self reports                 & C (7)           & No         & No                            \\
SemEval    & tweets                     & C (11)          & Yes        & No                            \\ GoEmotions
 & reddits                    & C (28)          & Yes        & Yes                           \\
Emobank    & multi-genre                & D (3)           & N/A        & Yes                           \\ \bottomrule
\end{tabular}
\vspace{-5pt}
\end{table}

\textbf{International Survey on Emotion Antecedents and Reactions (ISEAR)}~\cite{wallbott1986universal} is an outcome of a psychological study aiming to understand the antecedents and reactions to seven basic emotions (joy, fear, anger, sadness, disgust, shame, guilt). It consists of 7.6k samples from firsthand emotional reports in text form. We randomly select 500 samples for our analysis.

\textbf{SemEval 2017 Task 4 (SemEval)}~\cite{rosenthal2019SemEval} is part of the International Workshop on Semantic Evaluation. It consists of Twitter text samples, each annotated with one or more of 11 emotion classes. Since this dataset is very unbalanced
, we conduct weighted sampling to select the 500 samples by applying log inverse frequency weighting to the labels, in order to include more emotion labels in our analysis. If a sample carries multiple emotions, the weighting is determined by the rarest label.

\textbf{GoEmotions}~\cite{demszky2020GoEmotions} is a comprehensive dataset with 58k samples derived from Reddit comments, designed for fine-grained emotion detection. It is characterized by its extensive range of 27 distinct emotion categories, including admiration, remorse, gratitude, etc. Each sample can be assigned one or more emotion labels, as well as an extra ``neutral'' option. We also apply log inverse frequency weighting in our selection of 500 samples, to address the label imbalance.

\textbf{Emobank}~\cite{Buechel2017-zn} consists of 10k English sentences balancing multiple genres (newspapers, blogs, etc.). The samples are annotated with dimensional emotion labels in the Valence-Arousal-Dominance (VAD) space on a 5-point scale. We focus on the valence score in this study, as it is most commonly included in related literature~\cite{Feng2023-pr, Zhang2023-gl}. Notably, EmoBank distinguishes between the emotional perceptions of writers and readers~\cite{buechel2017readers}; we use the reader's annotations, to be consistent with the perspective of GPT-4. 
We weight the samples by their log deviation from neutral score, to encourage the inclusion of stronger emotional content. I.e., $ w_i = log |V_i - 3 |$.

\section{Methods}
\subsection{GPT-4 Prompting}
\label{sec:prompt_method}
For each of the three emotion classification datasets, we collect two sets of GPT-4 annotations. In the first set of annotations, we ask GPT-4 to conduct classification annotations by making selections from a pre-determined set of emotion classes. Informed by the common prompting techniques detailed in Section \ref{sec:prompt_literature}, we follow an instruction-based prompting method, which is consistent with the tasks given to human annotators. We try to give GPT-4 similar instructions as those given to humans, based on the descriptions in the GoEmotions paper~\cite{demszky2020GoEmotions}. Additionally, we set up a persona in the beginning, which has been found to be effective in our preliminary experiments. 
As an example, the prompt we use for multi-label classification datasets (GoEmotions and SemEval) is shown below. 

\textit{``You are an emotionally-intelligent and empathetic agent. You will be given a piece of text, and your task is to identify all the emotions expressed by the writer of the text. You are only allowed to make selections from the following emotions, and don't use any other words: [Options]. Only select those ones for which you are reasonably confident that they are expressed in the text. If no emotion is clearly expressed, select `neutral'. Reply with only the list of emotions, separated by comma.''}

We make minimal modifications as needed for other tasks/datasets and all prompts we use across datasets/tasks can be found in our released code.

We then ask GPT-4 to freely generate descriptors of the expressed emotion, without a given range of options. We 
compare the generated descriptors with the classification results to understand how the granularity of emotion labels affect GPT-4's performance (Section \ref{result:zero-shot}). For Emobank, we use a similar prompt with the expected response being a integer number from 1 to 5, indicating the perceived valence of the expressed emotion. Using these prompts, we obtain GPT-4 emotion annotations on the 2000 samples selected from four datasets. We additionally obtain classification annotations on all of the GoEmotions dataset using GPT-4 for our model training analysis (Section \ref{result:training}).

\subsection{Automatic Evaluation Metrics}
\label{sec:auto_eval_metrics}
Following common approaches in previous work, we evaluate GPT-4's performance on two aspects: 1) agreement with human annotations~\cite{Feng2023-pr, Ding2023-xi}, and 2) potential to improve model performance when GPT-4's annotations are used as training data~\cite{Latif2023-wl, Zhang2023-gl} to train smaller models, in this case implemented by fine-tuning BERT. For classification, we use Unweighted Average Recall (UAR) and Macro-averaged F-1 (Macro-F1) scores as the metrics. UAR measures a model's ability to correctly identify instances of each class with equal importance, while Macro-F1 assesses the balance between precision and recall for all classes. For regression, we use Pearson Correlation Coefficient (PCC) to measure the strength and direction of the linear correlation, and Mean Absolute Error (MAE) to reflect the average error magnitude.

\vspace{-6pt}
\subsection{Supervised model: Finetuned BERT}
\vspace{-5pt}
\label{sec:bert}
We finetune BERT~\cite{devlin2018bert} models on the full training set of each dataset to serve as a supervised baseline. BERT is one of the most commonly used models for text classification tasks~\cite{gonzalez2020comparing} and has been used as a benchmark model for the GoEmotions dataset~\cite{demszky2020GoEmotions}. Besides, its smaller size means it can be run on a single GPU, so we also use it as our base model when comparing the training efficiency of different annotation sources. 

We use the same finetuning settings across all experiments: we use the ``bert-base-uncased'' model implemented in the transformers library, which has 110M parameters. We add a linear layer on top of the base model, and finetune the whole model on a training set with an AdamW optimizer and learning rate = 1e-5. We optimize a Binary Cross Entropy loss for multi-label classification tasks, a Cross Entropy loss for the single-label classification task, and a Mean Squared Error loss for the regression task. We train the model for 10 epochs, and use the model with best performance on a validation set for testing. For the regression task, we find the model not yet converged after 10 epochs, so we train the model for 30 epochs.
\vspace{-5pt}
\subsection{Human Evaluation}
\vspace{-5pt}
Human annotations often contain inaccuracies~\cite{ikediego2018crowd}, thus metrics based solely on human annotations can be biased. Therefore, we conduct a human evaluation study on samples where GPT-4 and the human evaluators do not agree, aiming to incorporate human perspectives into our evaluation.

We recruited four students from the University of Michigan as our evaluators, aged between 19 to 28 and including two females. They were presented with annotations from two sources (i.e., human vs. GPT-4 classification or GPT-4 classification vs. generation) without identification, and were asked to choose the one which they thought ``better and more accurately describes the emotion expressed in the text''. Each sample was evaluated by two evaluators, who were given an option to indicate uncertainty on each sample. For the classification tasks (ISEAR, SemEval and GoEmotions), we first found all samples that were annotated with disjoint sets of labels by the two sources. Note that we did not adjudicate annotations if they contained overlapping emotion labels as the differences can be subtle (e.g., ``anger'' vs. ``anger, annoyance''). The annotations were randomized and mixed from three different datasets to reduce the likelihood that 
evaluators could recognizing patterns associated with a specific source. For the regression task (Emobank), it is harder for evaluators to decide whether a given number is a more or less accurate valence rating for a given sample, especially when the ratings are close. Thus, we adopted a relative evaluation schema~\cite{metallinou2013annotation}. We found pairs in disagreement where one annotation source assigns sample A a significantly ($>1$ standard deviation) higher rating than sample B, while the other indicates reversed significance. We asked evaluators to indicate which of the two samples in each pair should have the higher valence. The order of the samples was randomized.


\vspace{-5pt}
\section{Results}
\begin{figure*}[th]
    \centering
    \begin{minipage}[b]{0.33\textwidth}
        \includegraphics[width=\textwidth]{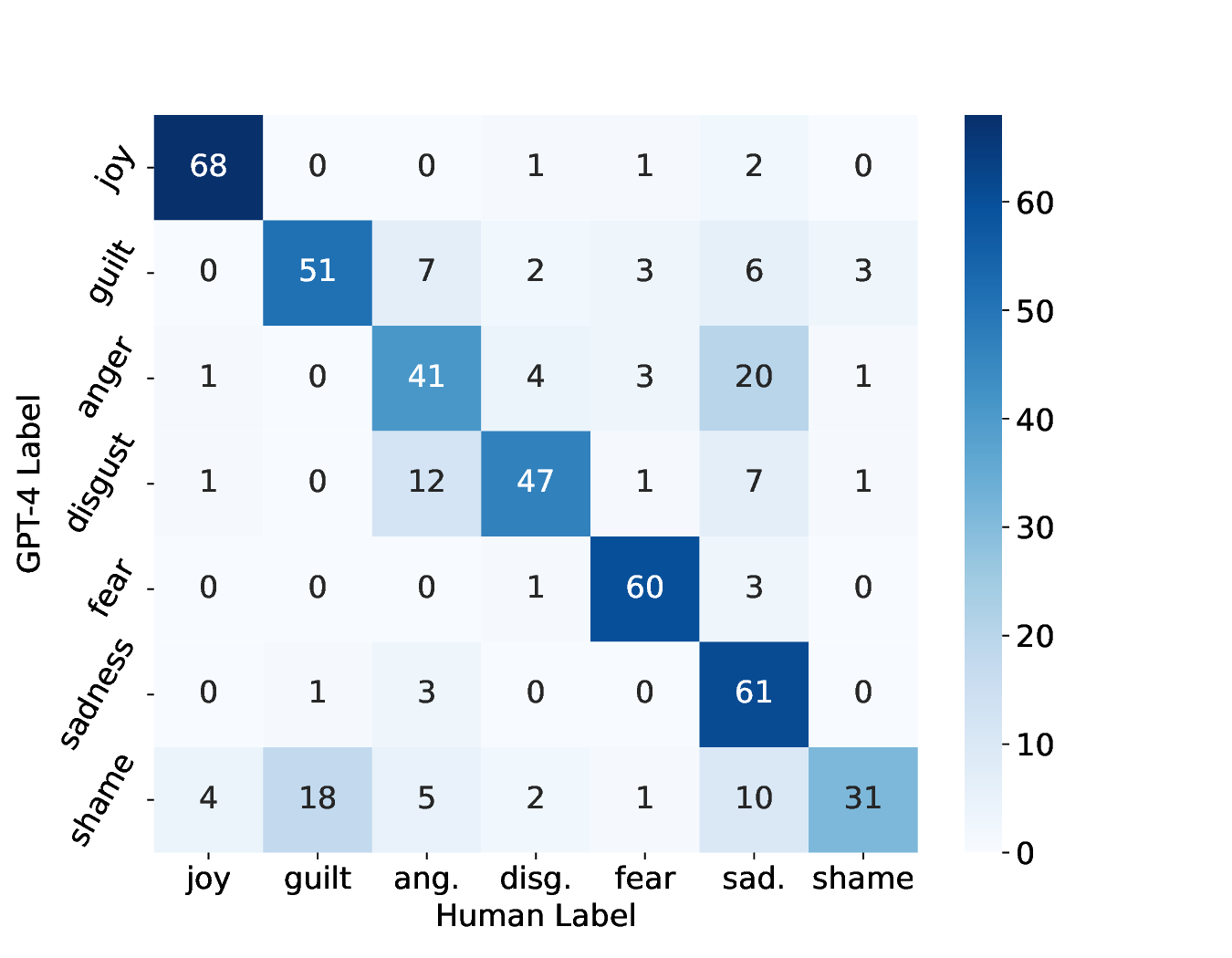}
        \caption{Disagreements between human and GPT annotations on ISEAR.}
        \label{fig:ISEAR}
    \end{minipage}
    \hfill 
    \begin{minipage}[b]{0.62\textwidth}
         \begin{minipage}[b]{0.5\linewidth}
            \centering
            \includegraphics[width=0.48\textwidth]{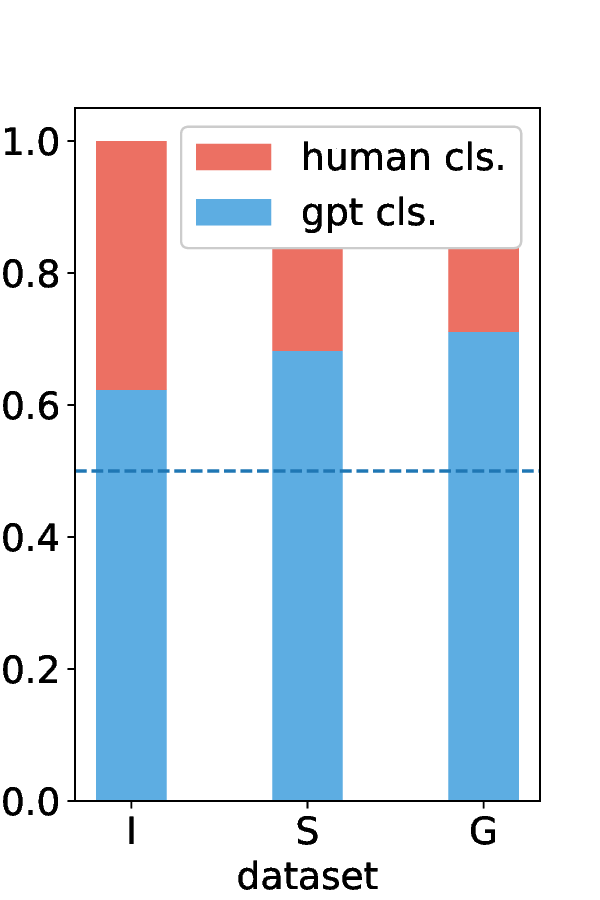}
            \hfill
            \includegraphics[width=0.48\textwidth]{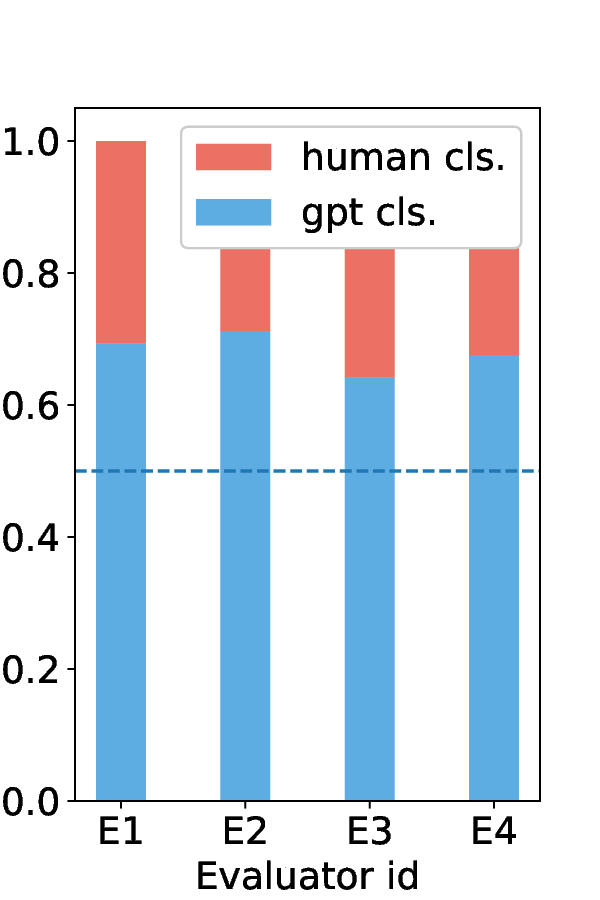}
            \subcaption{Human vs. GPT-4 classification.}
            \label{fig:first_pair}
        \end{minipage}
        \hfill
        \begin{minipage}[b]{0.5\linewidth}
            \centering
            \includegraphics[width=0.48\textwidth]{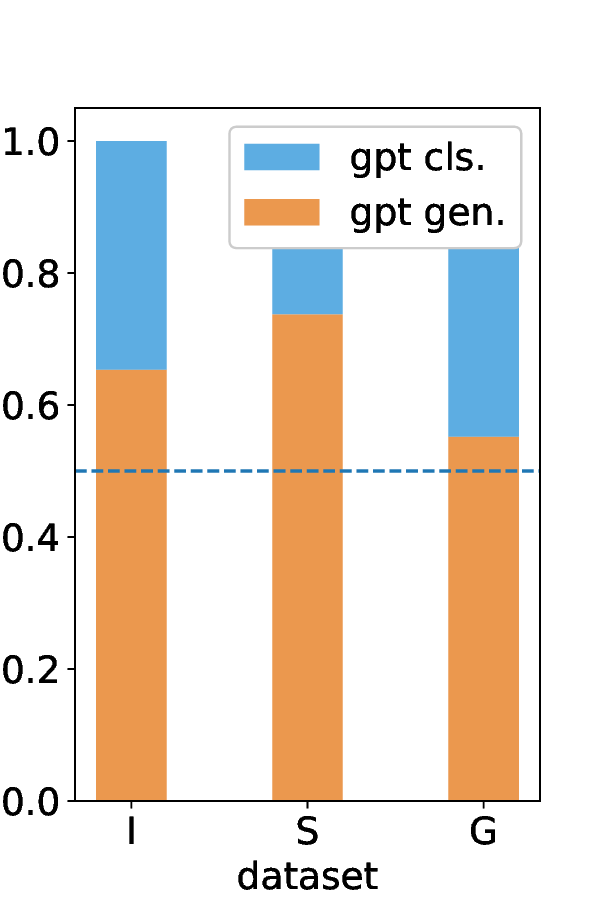}
            \hfill
            \includegraphics[width=0.48\textwidth]{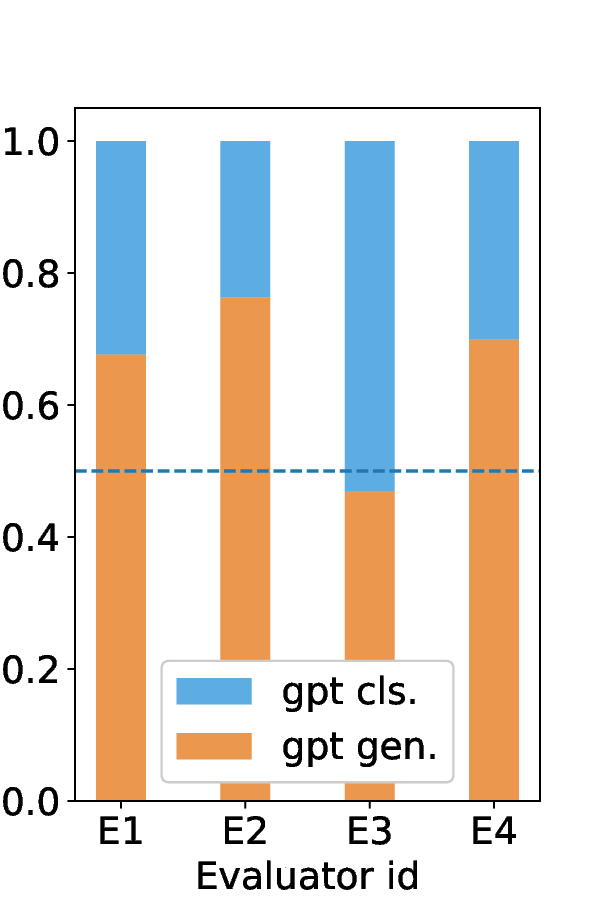}
            \subcaption{GPT-4 classification vs. generation.}
            \label{fig:second_pair}
        \end{minipage}
        \caption{Human preference ratio comparing human annotations, GPT-4 classification annotations and GPT-4 generation annotations on emotion classification tasks.}
        \label{fig:human_preference}
    \end{minipage}
    
\vspace{-5pt} 
\end{figure*}

\subsection{GPT-4 Zero-shot Performance}
\label{result:zero-shot}

We first compare GPT-4 and human classification annotations, with a focus on their disagreements. We visualize the disagreements on the ISEAR dataset as a confusion matrix in Figure \ref{fig:ISEAR} (we select ISEAR because it has the fewest number of classes and is the clearest to show). GPT-4 aligns well with human annotations on most samples, as indicated by the numbers on the diagonal. It's worth noting that confusions mostly happen among similar emotions, and the confusion between a positive emotion and a negative one is rare. Further, it shows that the confusion between classes is not symmetric, indicating some systematic differences between human and GPT-4 annotations. For example, GPT-4 tends to perceive more shame than guilt (18), but seldom marks human-perceived shame as guilt (3).

\begin{table}[t]
\centering
\vspace{-5pt}
\caption{GPT-4 zero-shot vs. BERT finetuned performance across four dataset. Better performances are in bold.}
\begin{tabular}{@{}lllll@{}}
\toprule
           & \multicolumn{2}{c}{\textbf{Macro-F1~~$\uparrow$}}                                            & \multicolumn{2}{c}{\textbf{UAR~~$\uparrow$}}                             \\ \midrule \vspace{-1pt}
           & \textit{GPT-4   }                           & \textit{BERT  }                             & \textit{GPT-4  }                            & \textit{BERT }          \\ \vspace{-1pt}
ISEAR      & \multicolumn{1}{l}{\textbf{0.739}} & 0.726                              & \multicolumn{1}{l}{\textbf{0.747}} & 0.727          \\ \vspace{-1pt}
SemEval    & \multicolumn{1}{l}{0.511}          & \textbf{0.548}                     & \multicolumn{1}{l}{0.476}          & \textbf{0.495} \\ \vspace{-1pt}
GoEmotions & \multicolumn{1}{l}{0.375}          & \multicolumn{1}{l}{\textbf{0.521}} & \multicolumn{1}{l}{\textbf{0.485}} & 0.469          \\ \midrule \vspace{-1pt}
           & \multicolumn{2}{c}{\textbf{PCC~~$\uparrow$}}                                                 & \multicolumn{2}{c}{\textbf{MAE~~$\downarrow$}}                             \\  \midrule \vspace{-1pt}
           & \textit{GPT-4 }                             & \textit{BERT }                              & \textit{GPT-4}                              & \textit{BERT}           \\ \vspace{-1pt} 
Emobank    & \textbf{0.764   }                           &       0.321                             & 0.645                              &   \textbf{ 0.442  }          \\  \bottomrule \vspace{-10pt} 
\end{tabular}
\label{table:zero-shot}
\vspace{-10pt} 
\end{table}
 
We then quantitatively evaluate the zero-shot efficacy of GPT-4, and compare its performance to a BERT model fine-tuned to predict the human evaluations. Our findings are in line with prior work in this space~\cite{Ding2023-xi} that the two approaches perform comparably, and GPT-4 performs slightly better than BERT on the easier 7-class classification dataset ISEAR, but was more challenged on the multi-label classification datasets SemEval and GoEmotions (Table \ref{table:zero-shot}). 

However, the subsequent human evaluation reveals a different trend and suggests that the automatic metrics may have underestimated GPT-4 performances. As shown in Figure~\ref{fig:human_preference} (a) with the colored bars representing the ratio of human preference obtained on each annotation source (Human vs. GPT-4 classification), human evaluators prefer labels from GPT-4 on more samples than those from human annotators, consistently across datasets: ISEAR 62.3\%, SemEval 68.2\%, GoEmotions 71.1\%. This trend holds for each individual annotator, ranging from 64.1\% to 71.2\%. 

Further, Figure~\ref{fig:human_preference} (b) shows that GPT-4 generated emotion descriptions are preferred to GPT-4 classification annotations by human evaluators, indicating that without the pre-defined classes as a restriction, GPT-4 generates emotion descriptions that were more often preferred by human evaluators. This trend is more significant when the label set is small, like ISEAR (7-classes 65.4\%) and SemEval (11 classes, 73.8\%), compared to GoEmotions (28 classes, 55.2\%). This comparison indicates that it's beneficial to have a larger label space, which is more likely to encompass the precise emotion labels needed for accurate annotation. The results in Figure~\ref{fig:human_preference} (a) highlight the proficiency of GPT-4 in navigating a wide range of labels, further demonstrating its utility in complex emotion recognition tasks. 

On the valence regression task, GPT-4 significantly outperforms fine-tuned BERT when measured by PCC, but it has a larger MAE (Table~\ref{table:zero-shot}). The large MAE can be explained by the highly centralized distribution of human annotations (standard deviation for human evaluations was 0.54, vs. 1.16 for GPT-4) and the fact that GPT-4 predicts integer-valued numbers while the human evaluations are continuous (e.g., averages of multiple evaluators). However, the large PCC value (0.764) indicates that GPT-4 can identify relative emotional valence. Human evaluation also finds an overall 56\% agreement with GPT-4 rather than the original human annotations.

\subsection{Impact on Model Training}
\label{result:training}
We then investigate whether the labels resulting from GPT-4 classification annotations can be used to train emotion recognition models. We focus on the GoEmotions dataset for this study, using its original train/val/test split.

We compare the performance of a BERT model when it is fine-tuned on the whole training set ($N_{train}$ = 42,278) with human annotations to one trained using the GPT-4 annotations. Additionally, we downsample the training data, retaining only data where the original human evaluations agree with the GPT-4 annotations.  We refer to this set as the filtered human set (Human-F, $N_{train}$ = 19,130).  
Note that this set potentially contains easier samples, compared to both the original human evaluations and GPT-4 annotation labels, because ambiguous samples are more likely to receive different annotations from human and GPT-4, and would thus be filtered out. We test the model on the original human evaluation data ($N_{test}$ = 5,283), the GPT-4 annotations ($N_{test}$ = 5,283), and the Human-F test set ($N_{test}$ = 2,409).  We add an extra test set that we refer to as the ``adjudicated'' test set ($N_{test} = 405$), which is a subset of the 500 samples used in the human preference evaluation experiment. The set is first populated with samples that have overlapping labels from the original human evaluations and GPT-4 ($N = 217$), and either the human or the GPT-4 label is selected by random. The remaining samples exhibit disagreement between the two sources.  We select the subset of samples where humans exhibited a clear preference for either the original human evaluation or GPT-4 label\footnote{Samples where the human evaluators did not agree on the preferred annotation were not included in this sample, 19\% of the samples.} as the final label ($N = 188$). The performance on the adjudicated test set is our main metric, because the annotations have been adjudicated and are considered to be more reliable than the raw human or GPT-4 annotations. 

\begin{table}[t]
\small
\centering
\setlength{\tabcolsep}{3pt}
\caption{Performance (Macro-F1) of models trained and tested on different combinations of annotations. We show the best performance on each test set (per column) in bold.}
\begin{tabular}{@{}c|cccc@{}}
\toprule
\diagbox{Train}{Test} & Human & GPT-4 & Human-F & Adjudicated \\ \midrule
\vspace{-1pt}
Human (42k)~~  & \textbf{0.486} & 0.304 & 0.568 & 0.392 \\ \midrule
\vspace{-1pt}
GPT-4 (42k)~~  & 0.343 & \textbf{0.517} & 0.533 & \textbf{0.524} \\ \midrule
\vspace{-1pt}
Human-F (19k)~~ & 0.478 & 0.367 & \textbf{0.617} & 0.430  \\ 
\bottomrule

\end{tabular}
\vspace{-5pt}
\label{bert-finetune-performance}
\end{table}

In Table~\ref{bert-finetune-performance},  the models perform most accurately when trained and tested on the same type of annotation. When models are trained on human annotations and tested with GPT-4 annotations (and vice versa) there are notable performance decreases. This indicates that the models learn a systematic difference between human and GPT-4 annotations, which echos our findings in Section \ref{result:zero-shot}.  
On the adjudicated test set, we find that the model trained on GPT-4 annotations outperformed the model trained on human annotations by a large margin (0.524 vs. 0.392, respectively), again pointing to the systematic differences between the two annotation sources. We find that models trained on the filtered subset of the original human evaluations estimate the labels of the adjudicated data more accurately than models trained on the full set of the original human evaluations (0.430 vs. 0.392, respectively). This result is notable given that the Human-F set is only 45\% of the size of the original Human set ($N=19,130$ vs. $N=42,278$, respectively). 


\section{Discussion, Limitations and Conclusion}
In this work, we evaluate GPT-4's emotion recognition capability and find that its zero-shot performance is comparable to supervised models. Our human evaluation study reveals that GPT-4 annotations are preferred to human annotations by our human evaluators, and GPT-4 is good at handling a wide range of options in emotion classification tasks. We also show that models trained on GPT-4 annotations are subsequently better at predicting the labels amongst the adjudicated subset of data. These results highlight the potential of LLMs to be applied in emotion recognition applications.

Several factors may contribute to the observed preference for GPT-4 annotations. First, humans make mistakes, and the increased cognitive load on more complex label spaces could have increased the vulnerability~\cite{wood2018comparison}. Additionally, given the inherent subjectivity and ambiguity of emotion annotations~\cite{devillers2005challenges}, different preferences could indicate variations in annotation perspectives or reflect a lack of diversity in the annotation process. Further exploration is needed to identify the underlying reason. Our findings emphasize the need to reconsider conventional notions of ``ground truth'' and explore novel evaluation metrics as LLMs approach and surpass human-level performance.

\bibliographystyle{IEEEtran}
\bibliography{mybib}

\end{document}